\documentclass{article} % For LaTeX2e
\usepackage{iclr2017_conference,times}
\usepackage{hyperref}
\usepackage{url}
\usepackage{algorithm,algpseudocode}
\usepackage{graphicx} % more modern
\usepackage{subfigure} 

\title{Variational Intrinsic Control}

\author{Karol~Gregor, Danilo Rezende and Daan Wierstra\\
DeepMind\\
\texttt{\{karolg,danilor,wierstra\}@google.com} \\
}

\begin{document}

\maketitle

\begin{abstract}
In this paper we introduce a new unsupervised reinforcement learning method for discovering the set of intrinsic options available to an agent. This set is learned by maximizing the number of different states an agent can reliably reach, as measured by the mutual information between the set of options and option termination states. To this end, we instantiate two policy gradient based algorithms, one that creates an explicit embedding space of options and one that represents options implicitly. The algorithms also provide an explicit measure of empowerment in a given state that can be used by an empowerment maximizing agent. The algorithm scales well with function approximation and we demonstrate the applicability of the algorithm on a range of tasks.
\end{abstract}

\section{Introduction}

In this paper we aim to provide an answer to the question what \emph{intrinsic} options are available to an agent in a given state -- that is, options that meaningfully affect the world. We define options as policies with a termination condition, and we are primarily concerned with their consequences -- what states in the environment they reach upon termination. The set of all options available to an agent is independent of an agent's intentions -- it is the set of all things that are possible for an agent to achieve. The purpose of this work is to provide an algorithm that aims to discover as many intrinsic options as it can, using an information theoretic learning criterion and training procedure.

This differs from the traditional approach to option learning where the goal is to find a small number of options that are useful for a particular task \citep{sutton1999between,mcgovern2001automatic,stolle2002learning,silver2012compositional,kulkarni2016hierarchical,mankowitz2014time,vezhnevets2016strategic,bacon2015option}. Limiting oneself to working with relatively small option spaces makes both credit assignment and planning over long time intervals easier. However, we argue that operating on the larger space of intrinsic options, as alluded to above, is in fact useful even though the space is vastly larger. First, the number of options is still much smaller than the number of all action sequences, since options are distinguished in terms of their final states, and many action sequences can reach the same state. Second, we aim to learn good representational embeddings of these options, where similar options are close in representational space and where we can rely on the power of generalization. In such embedded spaces a planner needs only choose a neighborhood of this space containing options that have sufficiently similar consequences. 

The idea of goal and state embeddings, along with a universal value function for reaching these goals, was introduced in \cite{schaul2015universal}. This work allowed an agent to efficiently represent control over many goals and to generalize to new goals. However, the goals were assumed to be given. This paper extends that work and provides a mechanism for learning goals (options) while preserving their embedded nature. 

There are at least two scenarios where our algorithm can be useful. One is the classical reinforcement learning case that aims to maximize an externally provided reward, as we explained above. In this case, rather than learning options to uniformly represent control, the agent can  combine extrinsic reward with an intrinsic control maximization objective, biasing learning towards high reward options. 

The second scenario is that in which the long-term goal of the agent is to get to a state with a maximal set of available intrinsic options -- the objective of \emph{empowerment} \citep{salge2014empowerment}. This set of options consists of those that the agent knows how to use. Note that this is not the theoretical set of all options: it is of no use to the agent that it is possible to do something if it is unable to learn how to do it. 
Thus, to maximize empowerment, the agent needs to simultaneously learn how to control the environment as well -- it needs to discover the options available to it. The agent should in fact not aim for states where it has the most control according to its current abilities, but for states where it expects it will achieve the most control \emph{after} learning. Being able to learn available options is thus fundamental to becoming empowered.

Let us compare this to the commonly used intrinsic motivation objective of maximizing the amount of model-learning progress, measured as the difference in compression of its experience before and after learning \citep{schmidhuber1991curious, schmidhuber2010formal, bellemare2016unifying, houthooft2016variational}. The empowerment objective differs from this in a fundamental manner: the primary goal is not to understand or predict the observations but to control the environment. This is an important point -- agents can often control an environment perfectly well without much \emph{understanding}, as exemplified by canonical model-free reinforcement learning algorithms \citep{sutton1998reinforcement}, where agents only model action-conditioned expected returns. Focusing on such understanding might significantly distract and impair the agent, as such reducing the control it achieves.

Our algorithm can be viewed as learning to represent the intrinsic control space of an agent. Developing this space should be seen as acquiring universal knowledge useful for accomplishing a multitude of different tasks, such as maximizing extrinsic or intrinsic reward (see \cite{oudeyer2008can} for an overview and useful references). This is analogous to unsupervised learning in data processing, where the goal is to find representations of data that are useful for other tasks. The crucial difference here, however, is that rather than simply finding representations, we learn explicit policies that an agent can choose to follow. Additionally, the algorithm explicitly estimates the amount of control it has in different states -- intuitively, the total number of reliably reachable states -- and can as such be used for an empowerment maximizing agent. 

A most common criterion for unsupervised learning is data likelihood. For a given data set, various algorithms can be compared based on this measure. No such commonly established measure exists for the comparison of unsupervised learning performance in agents. One of the primary difficulties is that in unsupervised learning the data is known, but in control, an agent exists in an environment and needs to act in it in order to discover what states and dynamics it contains. Nevertheless, we should be able to compare agents in terms of the amount of intrinsic control and empowerment they achieve in different states. Just like there are multiple methods and objectives for unsupervised learning \citep{Goodfellow-et-al-2016-Book}, we can devise multiple methods and objectives for unsupervised control. Data likelihood and empowerment are both information measures: likelihood measures the amount of information needed to describe data and empowerment measures the mutual information between action choices and final states. Therefore we suggest that what maximum likelihood is to unsupervised learning, mutual information between options and final states is to unsupervised control.

This information measure has been introduced in the empowerment literature before \citep{salge2014empowerment,klyubin2005empowerment} along with methods for measuring it (such as \cite{blahut1972computation,arimoto1972algorithm}). Recently, \cite{mohamed2015variational} proposed an algorithm that can utilize function approximation and deep learning techniques to operate in high-dimensional environments. However, this algorithm considers the mutual information between sequences of actions  and final states. This corresponds to maximizing the empowerment over \emph{open loop} options, where the agent \emph{a priori} decides on a sequence of actions in advance, and then follows these regardless of (potentially stochastic) environment dynamics. Obviously this often limits performance severely as the agent cannot properly react to the environment, and it tends to lead to a significant underestimation of empowerment. In this paper we provide a new perspective on this measure, and instantiate two novel algorithms that use \emph{closed loop} options where actions are conditioned on state. We show, on a number of tasks, that we can use these to both significantly increase intrinsic control and improve the estimation of empowerment. 

The paper is structured as follows. First we formally introduce the notion of intrinsic control and its derivation from the mutual information principle. After that, we describe our algorithm for intrinsic control with explicit options and argue for its viability in the experimental section. Last, we touch on intrinsic control with implicit options, and demonstrate that it scales up even better. We conclude with a short discussion on the merits of the approach and possible extensions.

\section{Intrinsic Control and the Mutual Information Principle}

In this section we explain how we represent intrinsic options and the corresponding objective we optimize. 

We define an option as an element $\Omega$ of a space and an associated policy $\pi(a|s,\Omega)$ that chooses an action $a$ in a state $s$ when following $\Omega$. The policy $\pi$ has a special \emph{termination} action that terminates the option and yields a final state $s_f$. Now let us consider the following example spaces for $\Omega$. 1) $\Omega$ takes a finite number of values $\Omega \in \{1, \ldots, n\}$. This is the simplest case in which for each $i$ a separate policy $\pi_i$ is followed. 2) $\Omega$ is a binary vector of length $n$. This captures a combinatorial number $2^n$ of possibilities. 3) $\Omega \in R^d$ is a d-dimensional real vector. Here the space of options is infinite. It is expected that policies for nearby $\Omega$s will be similar in practice. 

We need to express the knowledge about which regions of option space to consider. Imagine we start in a state $s_0$ and follow an option $\Omega$. As environments and policies are typically stochastic, we might terminate at different final states at different times. The policy thus defines a probability distribution $p^J(s_f|s_0,\Omega)$.
%where the $J$ stands for `jumpy'. 
Now consider two different options. If they lead to very similar states, they should inherently, intrinsically, not be seen as different from one another. So how do we express our knowledge regarding the effective intrinsic option set in a given state?

To help answer this question, consider an example of a discrete option case with three options $\Omega_1, \Omega_2 ,\Omega_3$. Assume that $\Omega_1$ always leads to a state $s_1$ while both $\Omega_2$ and $\Omega_3$ always lead to a state $s_2$. Then we would like to say that we really have two \emph{intrinsic} options: $\Omega_1$ and $(\Omega_2,\Omega_3)$. If we were to sample these options in order to maximize  behavior diversity we would half of the time choose $\Omega_1$  and half of the time any one of $\Omega_2, \Omega_3$. The relative choice frequencies of $\Omega_2$ and $\Omega_3$ do not matter in this example. We express these choices by a probability distribution $p^C(\Omega|s_0)$ which we call the controllability distribution.

Intuitively, to maximize intrinsic control we should choose $\Omega$s that maximize the diversity of final states while, for given $\Omega$, controlling as precisely as possible what the ensuing final states are. The former can be expressed mathematically as entropy $H(s_f) = -\sum_{s_f} p(s_f|s_0) \log p(s_f|s_0)$ where $p(s_f|s_0) = \sum_\Omega p^J(s_f|s_0,\Omega)p^C(\Omega|s_0)$. The latter, for a given $\Omega$, can be expressed as the negative log probability $-\log p^J(s_f|s_0,\Omega)$ (the number of bits needed to specify the final state given $\Omega$) which then needs to be averaged over $\Omega$ and $s_f$. Subtracting these two quantities yields the objective we wish to optimize -- the mutual information $I(\Omega,s_f|s_0)$ between options and final states under probability distribution $p(\Omega,s_f|s_0) = p^J(s_f|s_0,\Omega)p^C(\Omega|s_0)$:
\begin{eqnarray}
I(\Omega, s_f|s_0) \hspace{-3mm} &=& \hspace{-3mm} -\sum_{s_f} p(s_f|s_0) \log p(s_f|s_0) + \sum_{\Omega,s_f} p^J(s_f|s_0,\Omega)p^C(\Omega|s_0) \log p^J(s_f|s_0,\Omega),\\
&=& \hspace{-2mm} -\sum_{\Omega} p^C(\Omega|s_0) \log p^C(\Omega|s_0) + \sum_{\Omega,s_f} p^J(s_f|s_0,\Omega)p^C(\Omega|s_0)\log p(\Omega|s_0,s_f).
\label{eq:I2}
\end{eqnarray}
The mutual information is symmetric and the second line contains its reverse expression. This expression has a very intuitive interpretation associated with it: we should be able to tell options apart if we can infer them from final states. That is, if for two options $\Omega_1$ and $\Omega_2$, upon reaching state $s_{f1}$, we can infer it was option $\Omega_1$ that was executed rather than $\Omega_2$, and when reaching a state $s_{f2}$ we can infer it was option $\Omega_2$ rather than $\Omega_1$, then $\Omega_1$ and $\Omega_2$ can be said to be \emph{intrinsically} different options. We would like to maximize the set of options -- achieve a large entropy of $p(\Omega|s_0)$ (the first term of (\ref{eq:I2})).  At the same time we wish to make sure these options achieve intrinsically different goals -- that is, that they can be inferred from their final states. This entails maximizing $\log p(\Omega|s_0,s_f)$, the average of which is the second term of (\ref{eq:I2}). 

The advantage of this formulation is the absence of the term $p(s_f|s_0)$ in the formulation, which is difficult to obtain as we would have to integrate over $\Omega$. In rewriting the derivation, however, the term $p(\Omega|s_0,s_f)$ was introduced, which we arrived at from $p^J(s_f|s_0,\Omega)p^C(\Omega|s_0)$ using Bayes' rule. The quantity $p^J(s_f|s_0,\Omega)$ is inherent to the environment, but obtaining Bayes' reverse $p(\Omega|s_0,s_f)$ is difficult. However, it has an interpretation as a prediction of $\Omega$ from final state $s_f$. It would be fortuitous if we could train a separate function approximator to infer this quantity. Fortunately this is exactly what the variational bound \citep{mohamed2015variational} provides (see Appendix 1 for derivation):

\begin{eqnarray}
I^{VB}(\Omega, s_f|s_0) = -\sum_{\Omega} p^C(\Omega|s_0) \log p^C(\Omega|s_0) + \sum_{\Omega,s_f} p^J(s_f|s_0,\Omega)p^C(\Omega|s_0)\log q(\Omega|s_0,s_f) %\leq I(\Omega, s_f|s_0),
\label{eq:VB}
\end{eqnarray}

where $q$ is an option inference function which can be an arbitrary distribution, and we have $I \geq I^{VB}$. In this paper we train both the parameters of $p^C(\Omega|s_0)$, $q(\Omega|s_0,s_f)$ and the parameters of  policy $\pi(a|s,\Omega)$  (which determines $p^J(s_f|s_0,\Omega)$) to maximize $I^{VB}$. 

\section{Intrinsic Control with Explicit Options}

In this section we provide a simple algorithm to maximize the variational bound introduced above. Throughout we assume we have distributions, policies, and other possible functions parameterized using recent function approximation techniques such as neural networks, and state representations are formed from observations using recurrent neural networks. However, we only calculate the mutual information between options and final observations instead of final states, and leave the latter for future work. Algorithm \ref{alg:MFOPMC} provides an outline of the basic training loop.

\begin{algorithm}
\begin{algorithmic}
\State Assume an agent in a state $s_0$
\For{episode = $1,M$} 
\State Sample $\Omega \sim p^C(\Omega|s_0)$
\State Follow policy $\pi(a|\Omega,s)$ till termination state $s_f$
\State Regress $q(\Omega|s_0,s_f)$ towards $\Omega$
\State Calculate intrinsic reward $r_I = \log q(\Omega|s_0,s_f) - \log p^C(\Omega|s_0)$
\State Use a reinforcement learning algorithm update for $\pi(a|\Omega,s)$ to maximize $r_I$.
\State Reinforce option prior $p^C(\Omega|s_0)$ based on $r_I$.
\State Set $s_0=s_f$
\EndFor
\State Note: Empowerment at $s$ is estimated by the reinforce baseline of $p^C$, which tracks $r_I$.
\end{algorithmic}
\caption{Intrinsic Control with Explicit Options}
\label{alg:MFOPMC}
\end{algorithm}

This algorithm is derived from (\ref{eq:VB}) in Appendix 2. Note again that $\pi$ appears in (\ref{eq:VB}) by determining the distribution of terminal states $p^J(s_f|s_0,\Omega)$. Here we give an intuitive explanation of the algorithm. In a state $s_0$ an agent tries an option $\Omega$ from its available options $p^C(\Omega|s_0)$. Its goal is to choose actions that lead to a state $s_f$ from which this $\Omega$ can be inferred as well as possible using option inference function $q(\Omega|s_0,s_f)$. If it can infer this option well, then it means that other options don't lead to this state very often, and therefore this option is intrinsically different from others. This goal is expressed as the intrinsic reward $r_I$ (discussed in the next paragraph). The agent can use any reinforcement learning algorithm \citep{sutton1998reinforcement}, such as policy gradients \citep{williams1992simple} or $Q$-learning \citep{watkins1989learning, werbos1977advanced}, to train a policy to maximize this reward. In the final state, it updates its option inference function $q$ towards the actual $\Omega$ chosen (by taking the gradient of $\log q(\Omega|s_0,s_f)$). It also reinforces the prior $p^C$ based on this reward -- if the reward were high, it should choose this option more often. Note that we can also keep prior $p^C$ fixed, for example to the uniform Gaussian distribution. Then, different values of $\Omega$ will result in different behavior through learning. 

The intrinsic reward $r_I$ equals, on average, the logarithm of the number of different options an agent has in a given state -- that is, the empowerment in that state. This follows from the definition of mutual information (\ref{eq:VB}) -- it is the expression we get when we take a sample of $\Omega$ and $s_f$. However, we also provide an intuitive explanation. The $\log p^C(\Omega|s_0)$ is essentially the negative logarithm of the number of different $\Omega$s we can choose (for the continuous case, imagine finely discretizing). However, not all $\Omega$s do different things. The region where $q(\Omega|s_0,s_f)$ is large defines a region of similar options. The empowerment essentially equals the number of such regions in the total region given by $p^C$. Taking the logarithm of the ratio of the total number of options $\sim 1/p^C$ to the number of options within a region $\sim 1/q$ gives us $\log q/p^C = \log q - \log p^C = r_I$. 

We train $p^C$ using policy gradients \citep{williams1992simple}. During training we estimate a baseline to lower the variance of weight updates (see Appendix 2). This baseline tracks the expected return in a given state -- intrinsic reward in this case, which equals the empowerment. As such, the algorithm actually yields an explicit empowerment estimate. 

\subsection{Experiments}

\subsubsection{Grid World \label{sec.gridworld}}

\begin{figure}[h!] %t
\vspace{-.0cm}
\begin{center}
\begin{minipage}{1.\textwidth}
\centering
\includegraphics[width=0.7\textwidth]{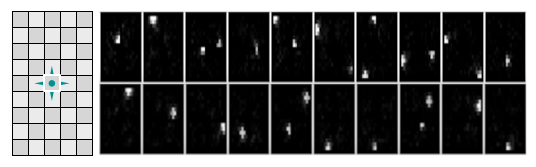}
\caption{\textbf{Learning to navigate through a grid world.} {\bf Left}: Standard grid world with valid actions indicated by green arrows (left, right, up, down) and by a green disc (do nothing). {\bf Right}: Each square corresponds to a different option, and shows the probability of predicting this option at different locations in the environment. The negative logarithm of these values is proportional to the intrinsic reward that a given option is trying to maximize, and therefore the locations with large intensity show the locations where a given option terminates. 
}\label{fig.gridworld}
%\vspace{-0.6cm}
% \label{fig:Grid}
\end{minipage}
\end{center}
\vspace{-.0cm}
\end{figure}

We demonstrate the behavior of the algorithm on a simple example of a two-dimensional grid world. The agent lives on a grid and has five actions -- it can move up, down, right, left and stay put. The environment is noisy in the following manner: after an agent takes a step, with probability $0.2$ the agent is pushed in a random direction. We follow Algorithm \ref{alg:MFOPMC}. We choose $\Omega$ to be a discrete space of $N=30$ options. We fix the option prior $p^C$ to be uniform (over $30$ values). The goal is therefore to learn a policy $\pi(a|s,\Omega)$ that would make the $30$ options end at as different states as possible. This is measured by the function $q(\Omega|s_f)$ which, from the state $s_f$ reached, tries to infer which option $\Omega$ was followed. At the end of an episode we get an intrinsic reward $r_I = -\log p + \log q = \log N + \log q$ ($\log N$ because $p^C = 1/N$ is fixed). If a particular option is inferred correctly and with confidence, then $\log q$ will be close to zero and negative, and the reward will be large ($\approx \log N$). If it is wrong, however, then $\log q$ will be very negative and the reward small. As we are choosing options at random (from the uniform $p^C$), in order to get a large reward on average, different options need to reach substantially different states in order for the $q$ to be able to infer the chosen option. In a grid world we can plot at which locations a given option is inferred by $q$, which are the locations to which the option navigates. This is shown in the Figure \ref{fig.gridworld} top, with each rectangle corresponding to a different option, and the intensity denoting the probability of predicting a given option. Thus, we see that indeed, different options learn to navigate to different, localized places in the environment. 

In this example, we use Q-learning to learn the policy. 
%(This paragraph is somewhat technical.) 
In general we can express the Q function for a set of $N$ different options by running the corresponding input states through a neural network and outputting $N \times n_{actions}$ values, one for each option and action. This way, we can update $Q$ of all the options at the same time efficiently, on a triplet of experience $s_t, a_{t}, s_{t+1}$. In this experiment we use a linear function approximator, and terminate options with fixed probability $1-\gamma = 0.05$. We could also use a universal value function approximation \citep{schaul2015universal}. For continuous option spaces we can still calculate the $Q$ by passing an input through a neural network, but then update the result on several, randomly sampled options $\Omega$ at the same time. Such an option space is then an option embedding in itself.

\begin{figure}[h!] %t
\vspace{-.0cm}
\begin{center}
\begin{minipage}{1.\textwidth}
\centering
\includegraphics[width=0.7\textwidth]{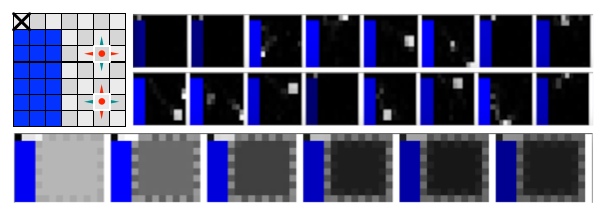}
\caption{\textbf{Learning to navigate through a grid world with trap states.} {\bf Left}: Dangerous grid world where cells on different checkerboard sub-lattices have different sets of valid actions (green arrows). Taking the wrong actions (red arrows and red disc) teleports the agent to the top-left corner indicated by a cross, from which there is only a small probability of escaping at each time step. The blue region represents a barrier, forming a narrow corridor on the top left. Furthermore, when a valid action is taken, with some probability the agent actually does not move. Therefore if the agent commits to a sequence of actions, it will quickly lose track as to which sub-lattice it is on, and fall into the low empowerment top left corner state. On the other hand, if it observes the environment it can always take actions to safely navigate the square. 
{\bf Bottom:} Classical empowerment, which considers the effect of (open loop) action sequences, has very small empowerment inside the square because each sequence does not respond to changes due to environment noise and the agent instead prefers to sit in the `safe' top left corridor. This is demonstrated in the bottom figure, which shows the exact empowerment at different locations for action sequence lengths $1,\ldots,6$. We see that not only does the square area have low empowerment, but that the amount of empowerment goes down with increasing sequence length. With closed loop policies, on the other hand, the agent has learned to navigate to different locations of the square, as shown in the top right figure.
}\label{fig.dangerous_gridworld}
%\vspace{-0.6cm}
% \label{fig:Grid}
\end{minipage}
\end{center}
\vspace{-.0cm}
\end{figure}

\subsubsection{'Dangerous' Grid World \label{sec.dangerous_gridworld}}

The second environment is also a grid world, but with special properties. It consists of two parts: a narrow corridor connected to an open square (see figure \ref{fig.dangerous_gridworld}, top-left), blue denoting the walls. However the square is not just a simple grid world, but is somewhat \emph{dangerous}. There are two types of cells arranged as a checkerboard lattice (as on a chess board). On one sub-lattice, only the left and right actions actually move the agent to the adjacent states and on the other sub-lattice only the up and down actions do. If the agent picks an action that is not one of these, it falls into a state where it is stuck for a long time. Furthermore, the move to an adjacent state only happens with some probability. Because of this, if the agent doesn't observe the environment, it quickly loses the information about which sub lattice it is on, and thus inevitably falls to the low empowerment state.
To show this we computed the exact empowerment values at each location for an open-loop policy using the Blahut-Arimoto algorithm for horizons $1$ to $6$, the result is shown in figure \ref{fig.dangerous_gridworld}(bottom).
On the other hand, if the agent observes the environment, it knows which sub-lattice it is on and can always choose an action that doesn't let it fall. Thus it can safely navigate the square. In our experiments, the agent indeed accomplishes this, and it learns options inside the square (see figure \ref{fig.dangerous_gridworld}, top-right). 

\subsection{The Importance of Closed Loop Policies}

Classical empowerment \citep{salge2014empowerment,mohamed2015variational} maximizes mutual information between sequences of actions $A = a_1,\ldots,a_T$ and final states. That is, it maximizes the same objective function %s (\ref{eq:I2}),
(\ref{eq:VB}), but where $\Omega = A$. 
This corresponds to maximizing empowerment over the space of open loop options. That is, options where an agent first commits to a sequence of actions and then blindly follows this sequence regardless of what the environment does. In contrast, in a closed-loop option every action is conditioned on the current state.
We show that using open-loop options can lead to severe underestimation of empowerment in stochastic environments, resulting in agents that aim to reach low-empowerment states.

We demonstrate this effect in the 'dangerous grid world' environment, section \ref{sec.dangerous_gridworld}. When using open loop options of length $T$, an agent at the center of the environment would have exponentially growing probability of being reset as a function of the option length $T$, resulting in an estimation of empowerment that quickly decreases with the option length, having its highest value inside the corridor at the top-left corner as shown in Figure \ref{fig.gridworld} (bottom).
A consequence of this is that such an agent would prefer being inside the corridor at the top-left corner, away from the center of the grid world.

In great contrast to the open loop case, when using closed loop options the empowerment will \emph{grow quadratically with the option length}, resulting in agents that prefer staying at the center of the grid world.

While this example might seem contrived, it is actually quite ubiquitous in the real world. For example, we can navigate around a city, whether walking or driving, quite safely. If we instead committed to a sequence of actions ahead, we would almost certainly be run over by a car, or if driving, crash into another car. 
%We would choose a different place to live. 
The importance of the closed loop nature of policies is indeed well understood. What we have demonstrated here is that one should not use open loop policies even to measure empowerment.

\subsection{Advantages and Disadvantages}
\label{AdvDis}

The advantages of Algorithm \ref{alg:MFOPMC} are: 1) It is relatively simple, 2) it uses closed loop policies, 3) it can use general function approximation, 4) it is naturally formulated with combinatorial options spaces, both discrete and continuous and 5) it is model-free. 

The primary problem we found with this algorithm is that it is difficult to make it work in practice with function approximation. We suggest there might be two reasons for this. First, the intrinsic reward is noisy and changing as the agent learns. This makes it difficult for the policy to learn. The algorithm worked as specified in those simple environments above when we used linear function approximation and a small, finite number of options. However, it failed when neural networks were substituted. We still succeeded by fixing the intrinsic reward for a period of time while learning the policy and vice versa. However, replacing the small option space by a continuous one made training even more difficult and only some runs succeeded. These problems are related to those in deep reinforcement learning \citep{mnih2015human}, where in order to make $Q$ learning work well with function approximation, one needs to store a large number of experiences in memory and replay them. It is possible that more work in this direction would find good practices for training this algorithm with general function and distribution approximations. 

The second problem is exploration. If the agent encounters a new state, it should like to go there, because it might correspond to some new option it hasn't considered before and therefore increase its control. However, when it gets there, the option inference function $q$ has not learned it yet. It is likely inferring the incorrect option, therefore giving a low reward and therefore discouraging the agent from going there. While the overall objective is maximized when the agent has the most control, the algorithm has difficulty maximizing this objective because two functions -- the intrinsic reward and the policy -- have to match up. It does a good job of expressing what the options are in a region it is familiar with, but it seems to fail to push into new state regions. Hence, we introduce a new algorithm formulation in section \ref{secImplicit} to address these issues.

\section{Intrinsic Control with Implicit Options}
\label{secImplicit}

To address the learning difficulties of Algorithm \ref{alg:MFOPMC} we use the action space itself as the option space. 
This gives the inference function $q$ grounded targets which makes it easier to train. Having a sensible $q$ makes the policy easier to train. The elements of the algorithm are as follows. The controllability prior $p^C$ and policy $\pi$ in Algorithm \ref{alg:MFOPMC} simply become a policy, which we denote by $\pi^p(a_t|s^p_t)$. The $s^p_t$ is an internal state that is calculated from $(s^p_{t-1}, x_t, a_{t-1})$. In our implementation it is the state of a recurrent network. The $q$ function in Algorithm \ref{alg:MFOPMC} should infer the action choices made by $\pi^p$ knowing the final observation $x_f$ and thus becomes $q = \pi^q(a_t|s^q_t)$ where $s^q$ is its internal state calculated from $(s^q_{t-1}, x_t, a_{t-1}, x_f)$. The logarithm of the number of action choices at $t$ that are effectively different from each other -- that can be distinguished based on the observation of the final state $x_f$ -- is given by $r_{I,t} = \log \pi^q(a_t|s^q_t) - \log \pi^p (a_t|s^p_t)$. We now introduce an algorithm that will maximize the expected cumulative number of distinct actions by maximizing the intrinsic return $R_I = \sum_t r_{I,t}$ in algorithm \ref{alg:Implicit}.

In this setting, maximization of control is substantially simplified. Consider an experience $x_0, a_0, \ldots, x_f$ generated by some policy. The learning of $\pi^q$ is a supervised learning problem of inferring the action choices that led to $x_f$. Even the random action policy terminates at different states, and thus $\pi^q$ is able to train on such experiences, mimicking decisions that happen to lead to $x_f$. The $\pi^p$ can be thought of as choosing among those $\pi^q$ that lead to diverse states, which in turn makes $\pi^q$ learn from experiences generated by those policies. The ability of $\pi^q$ to train on any experience motivated us to add an exploratory update in Algorithm \ref{alg:Implicit}.

%In this setting, maximization of control is substantially simplified. This can be seen by considering an experience $x_0, a_0, \ldots, x_f$ generated by an arbitrary policy (e.g. a uniform policy). The action choices during this experience have lead to the final state $x_f$. 
%Maximizing the intrinsic return $R_I$ with respect to $\pi^q(a_t|s^q_t)$ simply results in a supervised learning problem where we maximize the log-likelihood of the sequence $x_0, a_0, \ldots, x_f$ with respect to the parameters of $\pi^q$.
%Conversely, this is equivalent to learning a policy that will try to reach the final state $x_f$, starting at the state $x_0$ by regressing on past experience.
%This can be implemented very efficiently using replay buffers.

%For a fixed policy $\pi^q$, the policy $\pi^p$ that maximizes $R_I$ will try to reach a diverse set of final states $x_f$, which in turn makes $\pi^q$ learn about those policies. The ability of $\pi^q$ to train on any experience motivated us to add an additional exploratory update in Algorithm \ref{alg:Implicit}.

\begin{algorithm}
\begin{algorithmic}
%\State Start in $s_0$. 
\State {\bf Full update}
\State Follow policy $\pi^p(a_t|s^p_t)$, $s^p_t=f^p(s^p_{t-1}, x_t, a_{t-1})$ resulting in experience $x_0, a_0, \ldots, x_f$. 
\State For each $t$, regress policy $\pi^q(a_t|s^q_t)$, $s^q_t=f^q(s^q_{t-1},x_t,a_{t-1},x_f)$ towards action $a_t$
\State Calculate intrinsic reward $r_I = \sum_t \log \pi^q(a_t|s^q_t) - \log \pi^p(a_t|s^p_t)$
\State Reinforce the policy $\pi^p$ with $r_I$. 
\State {\bf Exploratory update}
\State Follow policy $\pi^p(a_t|s^p_t)$, $s^p_t=f^p(s^p_{t-1}, x_t, a_{t-1})$ with exploration ($\epsilon$ or other) resulting in experience $x_0', a_0', \ldots, x_f'$. 
\State For each $t$, regress policy $\pi^q(a_t'|s^q_t)$, $s^q_t=f^q(s^q_{t-1},x_t',a_{t-1}',x_f')$ towards action $a_t'$
%\State {\bf Note 1:} $s^p$ and $s^q$ are the internal states of $\pi^p$ and $\pi^q$ respectively.
\State {\bf Note:} Empowerment is estimated by the reinforce baseline of $\pi^p$.
\end{algorithmic}
\caption{Intrinsic Control with Implicit Options}
\label{alg:Implicit}
\end{algorithm}

In the experiments that follow, we use the following functions for policies (see Appendix 3 for the equations). Every input $x_t$ is passed through a fully connected, one layer neural network with a standard rectifier non-linearity, resulting in an embedding $u(x_t)$. This is passed to an LSTM recurrent net \citep{hochreiter1997long} which outputs the policy probabilities $\pi^p$ over actions. For $\pi^q$, we concatenate the embedding $u(x_t)$ and $u(x_f)$ and pass through another one layer neural net to obtain a state $v(x_t,x_f)$. This, together with hidden state of the LSTM net, is passed to another LSTM network, which outputs the probabilities $\pi^q$ over actions. 

Two facts are worth highlighting here. First, Algorithm \ref{alg:Implicit} applies to general, partially observable environments since the policies $\pi^p$ and $\pi^q$ build their own internal states. However, more generally, we should use final states $s_f$ of the environment instead of observations $x_f$ as the set of states an agent can reach. We leave this aspect to future work. Second, the policy $\pi^p$ can be thought of as an implicit option. However, the embedding $u(x_f)$ of the final observation (or state more generally) can be thought of as an explicit option and the policy $\pi^q$ as the policy implementing this option.

\subsection{Experiments}

\begin{figure}[h!] %t
\vspace{-.0cm}
\begin{center}
\begin{minipage}{1.\textwidth}
%\includegraphics[width=0.6\textwidth]{fig/comprImages.png}
%\vspace{0.1cm}
\includegraphics[width=0.33\textwidth]{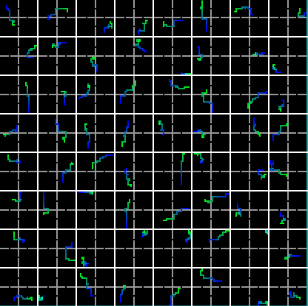}
\includegraphics[width=0.33\textwidth]{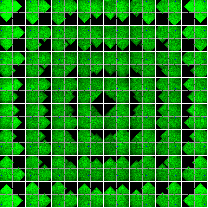}
\includegraphics[width=0.304\textwidth]{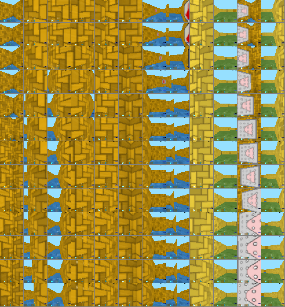}
%\vspace{0.1cm}
\caption{\textbf{Intrinsic Control with Implicit Options.} {\bf Left and Center}: The environment is a four room $25 \times 25$ grid world. Left: Blue denotes the beginning of a trajectory and green the end. The algorithm learns trajectories that extend through the environment and accurately pass through the doors between rooms. Center: Green shows the distribution of end points for trajectories of length 25. A trajectory in a given square starts at a location $(3x+1, 3y+1)$ where $(x,y)$ denotes the coordinates of the room in the picture. We see that the end points cover the reachable set of points nearly uniformly. {\bf Right}: Each column shows a trajectory in a three dimensional simulated environment. The agent only observes images. Again, we obtain very non-random trajectories that rotate or move by a large amount. The average amount of intrinsic control achieved is $5.4$ nats, corresponding to $\exp(5.4)=221$ different states.
}
\vspace{-0cm}
\label{fig:Implicit}
\end{minipage}
\end{center}
\vspace{-.0cm}
\end{figure}

We test this algorithm on several environments. The first one is a grid world of size $25 \times 25$ with four rooms (see Figure \ref{fig:Implicit} left (no action noise)). A random action policy of length $T$ leads to final states whose distance from the initial state is distributed approximately according to a Gaussian distribution of width $\sim \sqrt{T}$ within a room. For $T$ on the order of the environment size, such an agent rarely crosses to a different room, because of the narrow doors between the rooms. Figure \ref{fig:Implicit} shows trajectories learned by Algorithm \ref{alg:Implicit}. We see that indeed they are extended, spanning large parts of the environment. Furthermore many trajectories cross to different rooms passing straight through the narrow doors without difficulty. This is interesting, because while the policy $\pi^q$ is conditioned on the final state, the policy $\pi^p$ that is actually followed was not given any notion of the final state explicitly. It implicitly learns to navigate through the doors to different parts of the environment. 

To maximize intrinsic control, the distribution of final points that $\pi^p$ reaches should be uniform among the points that are reachable from a given state. This is because we can tell every point from any other point equally well. Figure \ref{fig:Implicit} (center) shows the distribution of the end points reached by the algorithm for trajectories of length $25$, starting at different points in the environment. For example, the top left square shows the end point distribution for starting at the top left corner. We see that the distribution is indeed roughly uniform among the reachable points. The average empowerment after learning reaches $6.0$ nats which corresponds to $\exp(6.0) \approx 403$ different reachable states.

In the second experiment we use a three dimensional simulated environment (Figure \ref{fig:Implicit}, right). At a given time the agent sees a particular view of its environment which is a $40 \times 40$ color image. The figure shows example trajectories that the agent follows (moving downwards in the figure) using policy $\pi^p$. We see that the trajectories seem intuitively much more `consistent' than those a random action policy would produce. The average empowerment achieved for trajectories of length $12$ was $5.4$ nats which corresponds to reaching $\exp(5.4)=221$ different states. 

\begin{figure}[h!] %t
\vspace{-.0cm}
\begin{center}
\begin{minipage}{0.3\textwidth}
\vspace{0.2cm}
\includegraphics[width=1.1\textwidth]{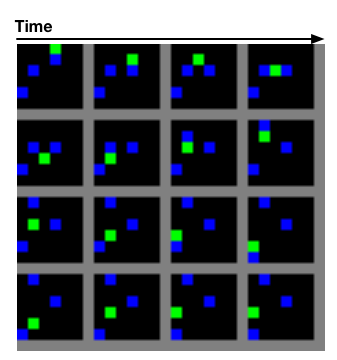}
\vspace{-0cm}
\end{minipage}
\hspace{1cm}
\begin{minipage}{0.5\textwidth}
\includegraphics[width=1.\textwidth]{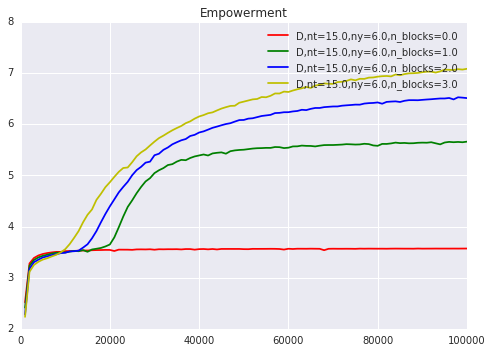}
\end{minipage}
\caption{ {\bf Pushing blocks.} {\bf Left:} The agent (green) lives in a $6 \times 6$ grid world and can push blocks (blue) around. The image shows a typical trajectory (read left to right). We see that the agent pushes blocks to various locations: it pushes the first block down, then it goes around the second one and pushes it up, the goes to the third one pushing it down, and then moves itself up to a final position. {\bf Right:} The empowerment as a function of training time in $6 \times 6$ world, with trajectories of length $15$ actions, with $0,1, 2, 3$ blocks. At the beginning of training, the agent first just learns it can go to different locations, but then it `realizes' that it can control the blocks. An average empowerment for three blocks reached $7.1$ nats which corresponds to being able to reach $\exp(7.1) \approx 1200$ states. }
\label{fig:Blocks}
\end{center}
\vspace{-.0cm}
\end{figure}

The third environment is again a grid world, but it contains blocks that the agent can push. The blocks cannot pass through each other or through the boundary, and the agent cannot push two blocks at the same time. In this case, the visual space is small, but there are combinatorially many possibilities. Figure \ref{fig:Blocks} shows a typical trajectory. We see that the agent pushes the first block down, then it goes around the second block and pushes it up, then goes to the third one and pushes it down, and then arrives at its final position. The average empowerment is $7.1$ nats which corresponds to being able to reach $\exp(7.1) \approx 1200$ different states.

\subsection{Elements Beyond an Agent's Control}

One prevalent feature of the real world is that there are many elements beyond our control such as falling leaves or traffic. One of the important features of these algorithms is that intrinsic options represent things that an agent can actually control and as such does not have to model all the complexities of the real world -- these algorithms are model-free. To demonstrate this property we introduce environments with elements beyond an agent's control. 

The first environment is the same four room grid world used above, but with two distractors moving around at random, that live on different input feature planes. These distractors do not affect the agent, but the agent does observe them. The agent needs to learn to ignore them. We find that the agent reaches the same amount of empowerment with and without the distractors (see Figure \ref{fig:distractors} in Appendix 4). 

\begin{figure}[h!] %t
\vspace{-.0cm}
\begin{center}
%\hspace{0.5cm}
\begin{minipage}{0.47\textwidth}  %43
\vspace{-0.2cm}
\includegraphics[width=1.05\textwidth]{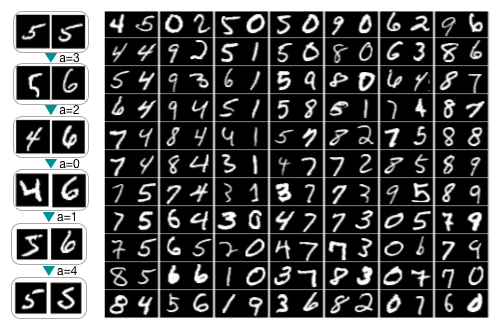}  %1.05
\end{minipage}
\hspace{0.35cm}  %0.3
\begin{minipage}{0.49\textwidth}  %49
\vspace{-0.2cm}
\includegraphics[width=1.\textwidth]{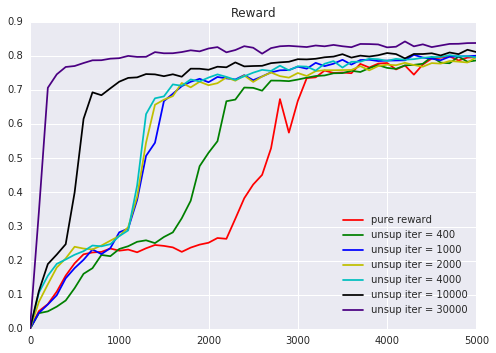}
\end{minipage}
%\vspace{0.1cm}
\caption{{\bf Left:} The environment consists of all possible pairs of MNIST digits. The agent can control the classes of the digits, being able to increase or decrease the class of the first or the second digit but the specific instance of a digit is sampled at random. Each two digit column of the figure shows a trajectory. A sample trajectory with the respective actions is shown on the left inset. The agent learns to navigate this space, achieving nearly maximum possible control of $100$ different states. Note that the agent does not see the digit class, it has to learn to tell different classes apart (to classify) through control. {\bf Right:} Extrinsic reward as a function training time for pure reinforce (red) and after a different amounts of unsupervised interaction with the environment. The environment is $15 \times 15$ four room grid world, and the reward is placed at location $(3,3)$. 
}
\vspace{-0.2cm}
\label{fig:beyondControl}
\end{center}
\vspace{-.0cm}
\end{figure}

The second environment consists of pairs of MNIST digits forming a $28 \times (2*28)$ image. There are five actions that affect classes of the digits. The first action doesn't change the classes, the next two increase/decrease the class of the first digit and the next two increase/decrease the class of the second digit (wrapping around). When a class is chosen, a random digit from that class is selected. Thus the environment is visually complex, but has a small control space. Example trajectories followed by the agent are shown in Figure \ref{fig:beyondControl}, left. The empowerment obtained by the agent with policies of length $10$ actions is $4.6$, which corresponds to $\exp(4.6)=99.5$ states. As there are 100 controllable states in the environment (but $60000^2$ total states), we see that agent achieves nearly maximum possible intrinsic control. 

\subsection{Open vs Closed Loop Options}
%Finally
Next we compare agents utilizing open and closed loop options. We use the grid world environment but add  noise as follows. After an agent takes a step, the environment pushes the agent in a random direction. While the closed loop policy can correct for the environment noise, for example by following a strategy of always going towards the goal, an open loop policy agent has less and less information on where it is in the environment as time goes by and it cannot navigate reliably towards the actual target location. We implement the open loop agent using the Algorithm \ref{alg:Implicit} but we only feed the starting and the end states to the recurrent network. Table \ref{table:open_closed} shows the comparison. We see that the closed loop agent indeed performs much better. 

\begin{table}[h]
\caption{\textbf{Open vs closed loop options}. Agents follow open or close loop options in a grid world with noise (see text for details). We show the empowerment as the exponential of the average number of nats, which approximately denotes how many different states can the agent reach.}
\begin{center}
\begin{tabular}{lcccc}
	\hline
	\textbf{Environment} & \textbf{Size} & \textbf{Option length} & \textbf{Open loop} & \textbf{Closed loop}  \\ 
        \hline
        Open grid & 6 & 6 & 2.7 & 5.8 \\
        Open grid & 6 & 12 & 2.8 & 6.7 \\
	Open grid & 10 & 10 & 3.7 & 12.6 \\ 
	Open grid & 10 & 20 & 4.1 & 15.1 \\ 
	Four room & 9 & 9 & 2.4 & 8.4 \\
	Four room & 15 & 15 & 3.3 & 19.0 \\
	Four room & 25 & 25 & 4.7 & 45.9 \\
   	\hline
\end{tabular}
\end{center}
\label{table:open_closed}
\end{table}

\subsection{Maximizing Extrinsic Reward}

Finally, while the primary focus of the paper is unsupervised control, we provide a proof-of-principle experiment that shows that learned policies can help in learning extrinsic reward. We consider the following situation. An agent is placed in an environment without being told what the objective is. The agent has an opportunity to explore and learn how to control the environment. After some amount of time, the agent is told the objective as an extrinsic reward $r_E$, and has a limited amount of time to collect as much reward as possible. There could be a number of ways to use the learned policies $\pi^p$ and $\pi^q$ to maximize the extrinsic reward. In our case we simply combine the intrinsic and extrinsic rewards, and reinforce the policy $\pi^p$ in Algorithm \ref{alg:Implicit} with $r = r_I + \alpha r_E$ where $\alpha$ is a large constant ($30$ in our experiment). We use the $15 \times 15$ four room grid world and, after different periods of time, we place a reward at location $3 \times 3$. Figure \ref{fig:beyondControl} (right) shows the reward collected per episode after different amounts of unsupervised pre-training or using pure reinforce without the maximum control objective (red curve). We see that, indeed, the agent learns to collect reward significantly faster after having an opportunity to interact with the environment.

\section{Conclusion}

In this paper we introduce a formalism of intrinsic control maximization for unsupervised option learning. We presented two algorithms in this framework and analyzed them in a diverse range of experiments. We demonstrated the importance of closed loop policies for estimating empowerment. Additionally, we also demonstrated the usefulness of unsupervised learning and intrinsic control for extrinsic reward maximization.

\subsubsection*{Acknowledgments}

We thank Shakir Mohamed, Frederic Besse, David Siver, Ivo Danihelka, Remi Munos, Ali Eslami, Tom Schaul, Nicolas Heess and Daniel Polani for useful discussions and comments.

%\bibliography{iclr2017_conference}
\bibliography{vmc}
\bibliographystyle{iclr2017_conference}

\section*{Appendix 1: Derivation of the Variational Bound on Mutual Information}
\label{ap:VB}

Here we derive the variational bound (\ref{eq:VB}). We drop the indices from variables and drop $s_0$: 

\begin{eqnarray}
I(\Omega, s) &=& -\sum_{\Omega} p(\Omega) \log p(\Omega) + \sum_{\Omega,s} p(\Omega, s)\log p(\Omega|s) \\
&=& -\sum_{\Omega} p(\Omega) \log p(\Omega) + \sum_{\Omega,s} p(\Omega, s)\log q(\Omega|s) + \sum_{s} p(s|\Omega) KL(p|q) \\
&\geq & -\sum_{\Omega} p(\Omega) \log p(\Omega) + \sum_{\Omega,s} p(\Omega, s)\log q(\Omega|s) = I^{VB}(\Omega, s).
\label{eq:VBderivation}
\end{eqnarray}

\section*{Appendix 2: Derivation of Algorithm \ref{alg:MFOPMC}}
\label{ap:MFOPMC}

Here we show the variational bound (\ref{eq:VB}) again, with parameters $\theta$ of $p^C$ and parameters $\phi$ of $q$ made explicit: 

\begin{eqnarray}
I^{VB}(\Omega, s_f|s_0) = -\sum_{\Omega} p_{\theta}^C(\Omega|s_0) \log p_{\theta}^C(\Omega|s_0) + \sum_{\Omega,s_f} p^J(s_f|s_0,\Omega)p_{\theta}^C(\Omega|s_0)\log q_{\phi}(\Omega|s_0,s_f) %\leq I(\Omega, s_f|s_0).
\end{eqnarray}

Taking the gradient with respect to $\phi$ is straightforward. We obtain a sample $\Omega \sim p_{\theta}^C(\Omega|s_0)$ and a sample of $s_f$ by following the policy till termination. Then we can directly take the gradient of $\log q$ which is the gradient of the standard likelihood maximization. 

Taking the gradient with respect to $\theta$ results in the following:

\begin{eqnarray}
\nabla_{\theta} I^{VB} &=& -\sum_{\Omega} p_{\theta}^C(\Omega|s_0) \left( - 1 - \log p_{\theta}^C(\Omega|s_0) +\sum_{s_f} p^J(s_f|s_0,\Omega) \log q_{\phi}(\Omega|s_0,s_f) \right) \nabla_{\theta} \log p_{\theta}^C(\Omega|s_0) \nonumber \\
&=& 
-\sum_{\Omega,s_f} p^J(s_f|s_0,\Omega) p_{\theta}^C(\Omega|s_0)  \left( - 1 - \log p_{\theta}^C(\Omega|s_0) +\log q_{\phi}(\Omega|s_0,s_f) \right) \nabla_{\theta} \log p_{\theta}^C(\Omega|s_0).
\end{eqnarray}

Now, we can add any number $b(s_0)$ inside the central bracket because the resulting sum is zero:
\begin{eqnarray}
&& \sum_{\Omega,s_f} p^J(s_f|s_0,\Omega) p_{\theta}^C(\Omega|s_0) b(s_0) \nabla_{\theta} \log p_{\theta}^C(\Omega|s_0) \\
&=& b(s_0) \sum_{\Omega,s_f} p^J(s_f|s_0,\Omega) \nabla_{\theta} p_{\theta}^C(\Omega|s_0) \\
&=& b(s_0) \nabla_{\theta} \sum_{\Omega} p_{\theta}^C(\Omega|s_0) \\
&=& b(s_0) \nabla_{\theta} 1 = 0.
\end{eqnarray}

Thus we obtain

\begin{eqnarray}
\nabla_{\theta} I^{VB} &=& -\sum_{\Omega,s_f} p^J(s_f|s_0,\Omega) p_{\theta}^C(\Omega|s_0)  \left( r_I - b(s_0) \right) \nabla_{\theta} \log p_{\theta}^C(\Omega|s_0) \\
r_I &=& \log q_{\phi}(\Omega|s_0,s_f) - \log p_{\theta}^C(\Omega|s_0).
\end{eqnarray}

This is the standard policy gradient update (along with its typical derivation): we sample $\Omega \sim p_{\theta}^C(\Omega|s_0)$, then pretend as we were training generative model $p^C$ on observed $\Omega$ by maximum likelihood (taking the gradient of $\log p^C$), but we multiply (reinforce) the resulting gradient by the corrected reward (return) $r_I - b$. The baseline $b$ aims to predict $r_I$ and is trained by regressing the values of $b$ towards observed $r_I$. Thus if in a given episode we obtain a  larger then expected reward, the likelihood of generating the same $\Omega$ is increased, and vice versa. In this particular case, because $r_I$ is a sampled empowerment at the state $s_0$, $b(s_0)$ tends towards the expected empowerment at that state. 

Finally we would like to update the policy. Keeping $\theta$ and $\phi$ fixed we would like to know how to change the policy to increase the $I^{VB}$. In a given sampled experience the value of $I^{VB}$ becomes $r_I$. We can use any reinforcement learning algorithm to maximize this reward. 

\section*{Appendix 3: Functional Forms for Intrinsic Control with Implicit Options}

We show the functional forms used in the Algorithm \ref{alg:Implicit} for the policies $\pi^p$ and $\pi^q$. Let $x_0,a_0,\ldots,x_f$ be a sequence of observations and actions. We assume we have computed embedding $u(x_f)$ of the final observation $x_f$. The computations at time step $t$ are the following:
\begin{eqnarray}
u_t &=& R W x_t, \\
h^p_t, c^p_t &=& \mbox{LSTM}([h^p_{t-1},u_t,a_{t-1}],c^p_{t-1}), \\
\pi^p_t &=& \mbox{softmax}(W h^p_t),\\
v_t &=& R W [u_t, u_f], \\
h^q_t, c^q_t &=& \mbox{LSTM}([h^q_{t-1},h^p_{t},v_t,a_{t-1}],c^q_{t-1}), \\
\pi^q_t &=& \mbox{softmax}(W h^q_t),
\end{eqnarray}
where $W$ is a linear operation (including biases that are suppressed from equations for clarity), $R$ is the rectifier nonlinearity, LSTM equations are below, $[a,b,\ldots]$ denotes concatenation of vectors $a,b,\ldots$ and $h$ and $c$ are states of LSTM.
The LSTM equations are
\begin{eqnarray}
h,c&=&\mbox{LSTM}(y,c) : \\
b_i,b_f,b_o,b_u &=& \mbox{split}(Wy), \\
g_i &=& \sigma(b_i),\\
g_f &=& \sigma(b_f),\\
g_o &=& \sigma(b_o),\\
u &=& \tanh(b_u),\\
c &=& g_f c + g_i u,\\
h &=& \tanh(c).
\end{eqnarray} 

\newpage
\section*{Appendix 4: The Effect of Distractors on Intrinsic Control}

\begin{figure}[t] %t
\vspace{-.0cm}
\begin{center}
\begin{minipage}{0.5\textwidth}
\includegraphics[width=1.\textwidth]{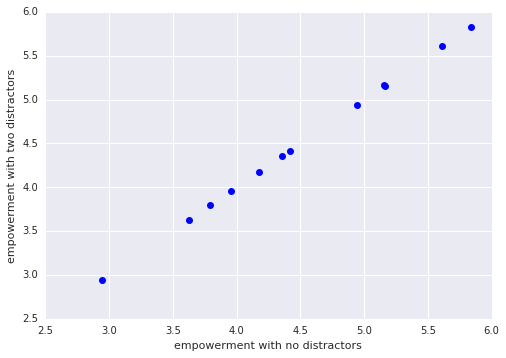}
\end{minipage}
%\vspace{0.1cm}
\caption{Empowerment in grid world with distractors vs grid world without them for different sizes and sequence lengths. Distractors move around the environment, the agents sees them but does not interact with them. The curve shows that the agent learns to completely ignore them and focus on the controllable part of the environment. 
}
\vspace{-0.2cm}
\label{fig:distractors}
\end{center}
\vspace{-.0cm}
\end{figure}

Figure \ref{fig:distractors} displays a graph showing that the agent learns to ignore distractors that do not affect its intrinsic control. The environment is a grid world with two points that move on a different feature plane and do not affect the agent, except as a visual distractor. We tested several environment sizes and option lengths. 

\end{document}